\documentclass[letterpaper, 10 pt, conference]{ieeeconf}  

\IEEEoverridecommandlockouts                              

\usepackage{cite}
\usepackage{amsmath,amssymb,amsfonts}

\usepackage{algorithmic}
\usepackage{algorithm}
\usepackage{graphicx}
\usepackage{textcomp}
\usepackage{xcolor}
\usepackage{booktabs}
\usepackage{multirow}
\usepackage{url}
\usepackage{hyperref}
\usepackage{bm}
\usepackage{mathtools}
\usepackage{siunitx}
\usepackage{cleveref}
\usepackage{cite}

\graphicspath{{../figures/}{../paper_figures/}}

\title{\LARGE \bf
Spatial-Semantic Uncertainty in VLM-Based Target Search: 
\\Balancing Exploration and Identification
}
\author{Alkesh Srivastava, Jonathan Diller, Vijay Kumar, Philip Dames
\thanks{Alkesh Srivastava and Philip Dames are with Temple University, Philadelphia, PA, 19122 USA. (e-mail: \{alkesh, pdames\}@temple.edu).}
\thanks{
Jonathan Diller and Vijay Kumar are with University of Pennsylvania, Philadelphia, PA, 19104 USA. (e-mail: \{diller, kumar\}@seas.upenn.edu)}
}

\begin{document}

\maketitle
\thispagestyle{empty}
\pagestyle{empty}

\begin{abstract}
Robots searching for a target from a natural-language description must determine not only \emph{where to search}, but also \emph{which observed candidate is the desired target}. These decisions reflect two distinct sources of uncertainty---spatial uncertainty over candidate locations and semantic uncertainty over target identity---that are often conflated in VLM-based search systems. We introduce a spatial--semantic uncertainty formulation that maintains separate beliefs over each component and integrates probabilistic VLM evidence into a global target-identity posterior, including probability mass for undiscovered targets. This decomposition allows an information-theoretic planner to independently value candidate discovery and target disambiguation through spatial and semantic expected information gain (EIG), providing an explicit mechanism for trading broader exploration against earlier identification. We evaluate six VLM uncertainty-elicitation interfaces on 500 synthetic targets and show that similar recognition accuracy can conceal substantial differences in calibration and false confidence. In degraded-observation search-and-identify experiments, EIG-based planners reach confident decisions in $75.0\%$--$92.5\%$ of trials, compared with $20.0\%$ for Random search, while different spatial--semantic weightings achieve comparable identification accuracy once confidence is attained. Increasing semantic emphasis reduces unnecessary exploration and VLM queries, demonstrating that explicitly planning over semantic uncertainty can accelerate target resolution without sacrificing decision quality. These results highlight the distinct roles of uncertainty representation and uncertainty-driven planning in embodied VLM systems.
\end{abstract}

\section{Introduction}
\label{sec:introduction}

\begin{figure}[t]
\centering
\includegraphics[width=0.75\linewidth]{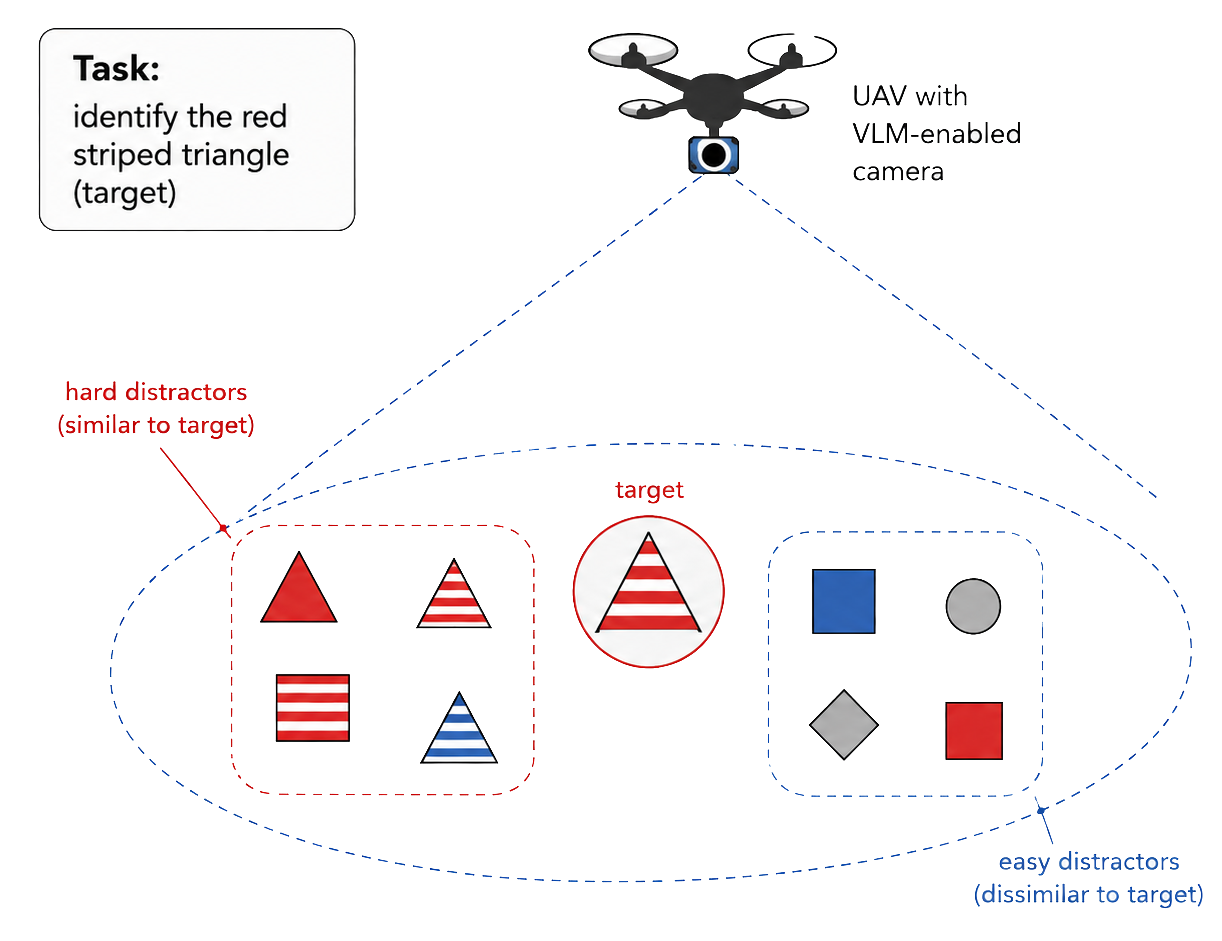}
\caption{VLM-based target search with visually similar distractors. The robot must discover candidate objects and resolve which candidate corresponds to the target.}
\label{fig:concept}
\end{figure}

Vision-language models (VLMs) increasingly enable robots to search for objects specified through natural-language descriptions. Such descriptions, however, may be incomplete or ambiguous, requiring a robot to distinguish the desired object from multiple visually plausible candidates \cite{majumdar2023findthis}. Moreover, VLM confidence is not necessarily a reliable indicator of correctness, and miscalibration can lead embodied agents to terminate exploration prematurely or continue searching unnecessarily \cite{ren2024explore,khan2024consistency}. As illustrated in~\Cref{fig:concept}, a robot searching for a ``red striped triangle'' may therefore encounter several plausible candidates before observing the true target. Successful search requires reasoning about two distinct questions: \emph{where should the robot continue searching?} and \emph{which observed candidate is the desired target?} These correspond to different sources of uncertainty and require different information-gathering actions. 

Probabilistic object search and semantic belief-space planning have long used spatial and semantic information to guide active sensing~\cite{tchuiev2023epistemic, kollar2009utilizing, liu2023active, srivastava2026bayesian}, while recent VLM-based systems incorporate semantic relevance or confidence into exploration and stopping decisions~\cite{ huang2026vlaconf, zhang2026igv}. In parallel, VLM uncertainty research has shown that confidence can be poorly calibrated and sensitive to the prediction or prompting interface \cite{khan2024consistency,schmalfuss2025parc}. However, existing work does not explicitly separate uncertainty over \emph{where candidate objects remain} from uncertainty over \emph{which candidate is the target}, nor study how weighting these two information sources changes the behavior of an active target-search system.
We address this gap with a spatial--semantic belief formulation that combines reliability-weighted probabilistic VLM evidence with a global target-identity posterior retaining probability mass for an undiscovered target. The decomposition exposes a direct trade-off between continued spatial exploration and early semantic resolution. Across a 500-target uncertainty study and full search-and-identify experiments under degraded observations, we find that VLM elicitation strategy strongly affects calibration despite similar recognition accuracy, while increasing semantic weight reduces explored candidates and VLM queries without materially changing identification accuracy once confidence is reached.

The contributions of this work are threefold: (i) a systematic comparison of six VLM uncertainty-elicitation interfaces, showing that similar recognition accuracy can conceal substantial differences in calibration and false confidence; (ii) a global spatial--semantic target belief that incorporates probabilistic VLM evidence while retaining an undiscovered-target hypothesis; and (iii) an empirical demonstration that decomposing spatial and semantic uncertainty gives the planner an explicit mechanism to trade candidate discovery against early target identification, with greater semantic emphasis reducing unnecessary exploration and VLM queries without degrading conditional identification accuracy.

\section{Problem Formulation}
\label{sec:problem_formulation}

Consider a robot operating in a bounded planar workspace $\mathcal{W}\subset\mathbb{R}^{2}$ and tasked with locating a particular static target $\mathcal{X}$ from a natural-language description $\mathcal{D}$, e.g., ``red barrel with a circular blue lid.'' The workspace contains a finite set of visually observable candidate objects, exactly one of which corresponds to $\mathcal{X}$. Since $\mathcal{D}$ may only partially specify the target appearance, multiple candidates may be visually consistent with the description.

Let $\mathcal{G}\subset\mathcal{W}$ denote the discrete search lattice. The
robot occupies a position \( \mathbf{r}_t=
\begin{bmatrix}
x_t & y_t
\end{bmatrix}^{\top}
\in\mathcal{G},
\) and moves on an $8$-connected grid according to
\begin{equation}
\mathbf{r}_{t+1}
=
\mathbf{r}_t+\mathbf{a}_t,
\qquad
\mathbf{a}_t\in\mathcal{A}_8,
\label{eq:grid_motion}
\end{equation}
where
\begin{equation}
\mathcal{A}_8
=
\left\{
\begin{bmatrix} i & j \end{bmatrix}^{\top}
\,\middle|\,
i,j\in\{-1,0,1\},\;
(i,j)\neq(0,0)
\right\},
\label{eq:eight_connected_actions}
\end{equation}
is the set of admissible $8$-connected motion primitives, restricted to moves such that $\mathbf{r}_{t+1}\in\mathcal{G}$.

Let $\mathcal{C}=\{c_{1},\ldots,c_{N}\}$ denote a set of stationary candidate objects in the workspace. While the total number of candidates $N$ is known \textit{a priori}, their locations $\mathbf{p}_{i} \in \mathcal{W}$ are unknown and discovered incrementally during the search. The target index is denoted by $c^{\star}\in\mathcal{C}$, such that candidate $c^{\star}$ corresponds to $\mathcal{X}$.

At each time step, the robot acquires a visual observation $\mathbf{z}_{t}$ from its current pose. Depending on the robot viewpoint and sensor field of view, the observation may reveal previously unseen candidates or provide additional visual evidence about candidates observed earlier. The natural-language description $\mathcal{D}$ is used to determine the semantic consistency between observed candidate appearances and the target specification.

Given the current robot position $\mathbf{r}_t$, the accumulated observation history $\mathcal{Z}_{1:t}$, and the target description $\mathcal{D}$, the robot selects a finite-horizon path
\(
\tau_t
=
\left(
\mathbf{r}_t,
\mathbf{r}_{t+1},
\ldots,
\mathbf{r}_{t+H}
\right),
\)
where each transition satisfies the $8$-connected motion constraint in~\eqref{eq:eight_connected_actions}. 

Let $\mathcal{T}(\mathbf{r}_t)$ denote the set of feasible paths originating from $\mathbf{r}_t$. We seek a search policy $\pi:(\mathbf{r}_t,\mathcal{Z}_{1:t},\mathcal{D})\mapsto\tau_t$ that minimizes the time required to reach a confident identification of the true target $c^\star$ under noisy observations, subject to the motion and sensing constraints. Achieving this objective requires balancing spatial exploration for candidate discovery against semantic information gathering for confident target identification.

\section{Spatial-Semantic Belief Representation}
\label{sec:belief_representation}

The search task in~\Cref{sec:problem_formulation} requires the robot to resolve uncertainty over both the spatial distribution of candidate objects and the identity of the true target $\mathcal{X}$. We therefore maintain separate spatial and semantic belief representations.

\subsection{Spatial Belief}
\label{sec:spatial_belief}

Let $\mathcal{G}={g_1,\ldots,g_{N_g}}\subset\mathcal{W}$ denote the discrete search lattice and define the latent occupancy map
\begin{equation}
\mathbf{M}
=
[M_1,\ldots,M_{N_g}]^\top,
\qquad
M_i\in\{0,1\},
\label{eq:latent_spatial_map}
\end{equation}
where $M_i=1$ indicates that cell $g_i$ contains a candidate object. We model the cells as independent Bernoulli random variables,
$M_i\sim\operatorname{Bernoulli}(p_i)$.

Let $C^\star$ denote the discrete random variable identifying which candidate corresponds to $\mathcal{X}$. Since candidate placement is independent of target identity,
\begin{equation}
p(\mathbf{M},C^\star)
=
p(\mathbf{M})p(C^\star),
\label{eq:spatial_semantic_factorization}
\end{equation}
which yields the additive decomposition
\begin{equation}
H(\mathbf{M},C^\star)
=
H(\mathbf{M})+H(C^\star).
\label{eq:spatial_semantic_decomposition}
\end{equation}
We denote the two components by
$\mathcal{H}^{\mathrm{sp}}\triangleq H(\mathbf{M})$
and
$\mathcal{H}^{\mathrm{sem}}\triangleq H(C^\star)$. Given accumulated spatial observations
$\mathcal{Z}^{\mathrm{sp}}_{1:t}$, let
$p_{i,t}=P(M_i=1\mid\mathcal{Z}^{\mathrm{sp}}_{1:t})$
denote the posterior occupancy probability of cell $g_i$. Under the independent-cell assumption, the spatial uncertainty is $\mathcal{H}^{\mathrm{sp}}_t
=
\sum_{i=1}^{N_g} h(p_{i,t})$, where 
$h(p)
=
-p\log p-(1-p)\log(1-p).$

\subsection{VLM Uncertainty Interface}
\label{sec:vlm_uncertainty_interface}

The semantic belief requires probabilistic visual evidence rather than a hard VLM prediction. We therefore use a Direct-Score interface in which the VLM assigns a nonnegative score to each possible value of a task-relevant semantic attribute. Let $\mathcal{A}$ denote the set of semantic attributes and $\mathcal{V}_a$ the categorical support of attribute $a\in\mathcal{A}$. For a discovered candidate $c_j$, the VLM assigns a score $s_t^{j,a}(v)\geq 0$ to each value $v\in\mathcal{V}_a$. These scores are normalized to obtain the categorical distribution
\begin{equation}
q_t^{j,a}(v)
=
\frac{s_t^{j,a}(v)}
{\sum_{v'\in\mathcal{V}_a}s_t^{j,a}(v')},
\qquad
\sum_{v\in\mathcal{V}_a}q_t^{j,a}(v)=1.
\label{eq:vlm_attribute_distribution}
\end{equation}

\subsection{Semantic Identity Belief}
\label{sec:semantic_identity_belief}

Let $\mathcal{C}_t=\{c_1,\ldots,c_{N_t}\}$ denote the candidates discovered by time $t$. Repeated observations of a candidate are fused in log-probability space. For candidate $c_j$, attribute $a$, and value $v$,
\begin{equation}
\Lambda_t^{j,a}(v)
=
\Lambda_{t-1}^{j,a}(v)
+
w_t^a\log q_t^{j,a}(v),
\label{eq:semantic_evidence_fusion}
\end{equation}
where $w_t^a\in[0,1]$ is an attribute-specific reliability weight obtained from a separate calibration procedure. The fused categorical belief is
\begin{equation}
p_t^{j,a}(v)
=
\frac{\exp(\Lambda_t^{j,a}(v))}
{\sum_{v'\in\mathcal{V}_a}\exp(\Lambda_t^{j,a}(v'))}.
\label{eq:fused_attribute_belief}
\end{equation}
Thus, observations with lower empirically measured reliability contribute proportionally less semantic evidence.

Let $\mathcal{A}_{\mathcal{D}}\subseteq\mathcal{A}$ denote the attributes specified by the target description $\mathcal{D}$, with desired value $d_a\in\mathcal{V}_a$ for each $a\in\mathcal{A}_{\mathcal{D}}$. Assuming conditional independence across described attributes, the semantic compatibility of candidate $c_j$ with $\mathcal{D}$ is
\begin{equation}
\mu_t^j
\triangleq
P(c_j\simeq\mathcal{D})
=
\prod_{a\in\mathcal{A}_{\mathcal{D}}}
p_t^{j,a}(d_a).
\label{eq:candidate_semantic_match}
\end{equation}

Because exactly one candidate corresponds to $\mathcal{X}$ and the true target may remain undiscovered, we construct a global identity posterior over
$\mathcal{C}_t\cup{\varnothing}$, where $\varnothing$ denotes the undiscovered hypothesis. Defining the candidate match odds
$o_t^j=\mu_t^j/(1-\mu_t^j)$, we obtain
\begin{align}
\pi_t(c_j)
&=
\frac{o_t^j}
{\sum_{k=1}^{N_t}o_t^k+(N-N_t)},
\label{eq:identity_posterior_candidate}\\
\pi_t(\varnothing)
&=
\frac{N-N_t}
{\sum_{k=1}^{N_t}o_t^k+(N-N_t)},
\label{eq:identity_posterior_undiscovered}
\end{align}
where $N$ is the total number of candidate objects in the environment.

The resulting semantic uncertainty is
\begin{equation}
\begin{aligned}
\mathcal{H}^{\mathrm{sem}}_t
={}&
-\sum_{c_j\in\mathcal{C}_t}
\pi_t(c_j)\log_2\pi_t(c_j) \\
&-\pi_t(\varnothing)\log_2\pi_t(\varnothing) \\
&+\pi_t(\varnothing)\log_2(N-N_t),
\end{aligned}
\label{eq:semantic_identity_entropy}
\end{equation}
where the final term preserves the uncertainty associated with the $N-N_t$ unresolved target hypotheses aggregated by $\varnothing$. When $N_t=N$, the undiscovered hypothesis is removed and the final two terms in~\eqref{eq:semantic_identity_entropy} vanish.

\section{Information-Theoretic Search and Identification}
\label{sec:information_planning}

We instantiate the search policy in~\Cref{sec:problem_formulation} using expected information gain over the spatial and semantic beliefs introduced in~\Cref{sec:belief_representation}. For each feasible path $\tau\in\mathcal{T}(\mathbf{r}_t)$, we compute the expected reduction in spatial and semantic uncertainty over a horizon $H$ as
\begin{align}
\mathcal{I}^{\mathrm{sp}}_t(\tau)
&=
\mathcal{H}^{\mathrm{sp}}_t
-
\mathbb{E}\!\left[
\mathcal{H}^{\mathrm{sp}}_{t+H}
\mid \tau
\right],
\label{eq:spatial_eig}\\
\mathcal{I}^{\mathrm{sem}}_t(\tau)
&=
\mathcal{H}^{\mathrm{sem}}_t
-
\mathbb{E}\!\left[
\mathcal{H}^{\mathrm{sem}}_{t+H}
\mid \tau
\right].
\label{eq:semantic_eig}
\end{align}

The selected path is then
\begin{equation}
\tau_t^\star
=
\arg\max_{\tau\in\mathcal{T}(\mathbf{r}_t)}
\left[
\alpha\,\mathcal{I}^{\mathrm{sp}}_t(\tau)
+
\beta\,\mathcal{I}^{\mathrm{sem}}_t(\tau)
\right],
\label{eq:path_planning_objective}
\end{equation}
where $\alpha,\beta\geq0$ determine the relative emphasis on spatial exploration and semantic target identification.

The spatial expectation is taken over future occupancy measurements along $\tau$. Since these measurements depend jointly on the unknown map and the stochastic observation process, we approximate the expectation using Monte Carlo rollouts. Let
$\mathbf{M}^{(k)}\sim p(\mathbf{M}\mid\mathcal{Z}^{\mathrm{sp}}_{1:t})$
denote the $k$th sampled map and
$\mathcal{Z}^{(k)}_{\tau}$
the corresponding simulated measurements along $\tau$. Then
\begin{equation}
\mathbb{E}\left[
\mathcal{H}^{\mathrm{sp}}_{t+H}
\mid\tau
\right]
\approx
\frac{1}{K}
\sum_{k=1}^{K}
\mathcal{H}^{\mathrm{sp}}
\left(
p(\mathbf{M}\mid
\mathcal{Z}^{\mathrm{sp}}_{1:t},
\mathcal{Z}^{(k)}_{\tau})
\right).
\label{eq:spatial_eig_mc}
\end{equation}

Semantic information gain is computed from the predictive identity posterior induced by candidate observations along $\tau$. This includes both first observations of newly discovered candidates and re-observations of previously discovered candidates for which the semantic belief remains uncertain. For a candidate $c_j$ expected to be observed, the prospective VLM outcome is marginalized under the calibrated semantic observation model, yielding the expected posterior entropy
\begin{equation}
\mathbb{E}\!\left[
\mathcal{H}^{\mathrm{sem}}_{t+H}
\mid\tau
\right]
=
\sum_{z\in\mathcal{Z}^{\mathrm{sem}}_{\tau}}
P(z\mid\tau)\,
\mathcal{H}^{\mathrm{sem}}
\left(
\pi_{t+H}^{z}
\right).
\label{eq:semantic_eig_expectation}
\end{equation}
where $\pi_{t+H}^{z}$ denotes the identity posterior obtained after hypothetical semantic observation $z$.

The robot continues search until the semantic posterior reaches a prescribed confidence threshold. We define the identification time as
\begin{equation}
T_{\mathrm{id}}
=
\inf
\left\{
t:
\max_{c_j\in\mathcal{C}_t}
\pi_t(c_j)
\geq\eta
\right\},
\label{eq:identification_time}
\end{equation}
where $\eta\in(0,1)$ is the confidence threshold. At termination, the robot declares
\begin{equation}
\hat{c}
=
\arg\max_{c_j\in\mathcal{C}_{T_{\mathrm{id}}}}
\pi_{T_{\mathrm{id}}}(c_j).
\label{eq:target_declaration}
\end{equation}
In our experiments, $\eta=0.95$. The stopping rule depends only on the robot's internal posterior and is therefore independent of the ground-truth target identity, which is used only for post-hoc evaluation of whether the declaration is correct.

\section{Experiments and Results}
\label{sec:experiments}

Using GPT-4o-mini as the vision-language model, we evaluate two questions:
(i) how probabilistic uncertainty should be elicited from the VLM, and
(ii) whether reasoning over spatial and semantic uncertainty improves
time-to-confidence of target identification under noisy observations.

\subsection{VLM Uncertainty Interface Study}
\label{sec:interface_experiment}

\begin{figure}[t]
\centering
\includegraphics[width=0.85\linewidth]{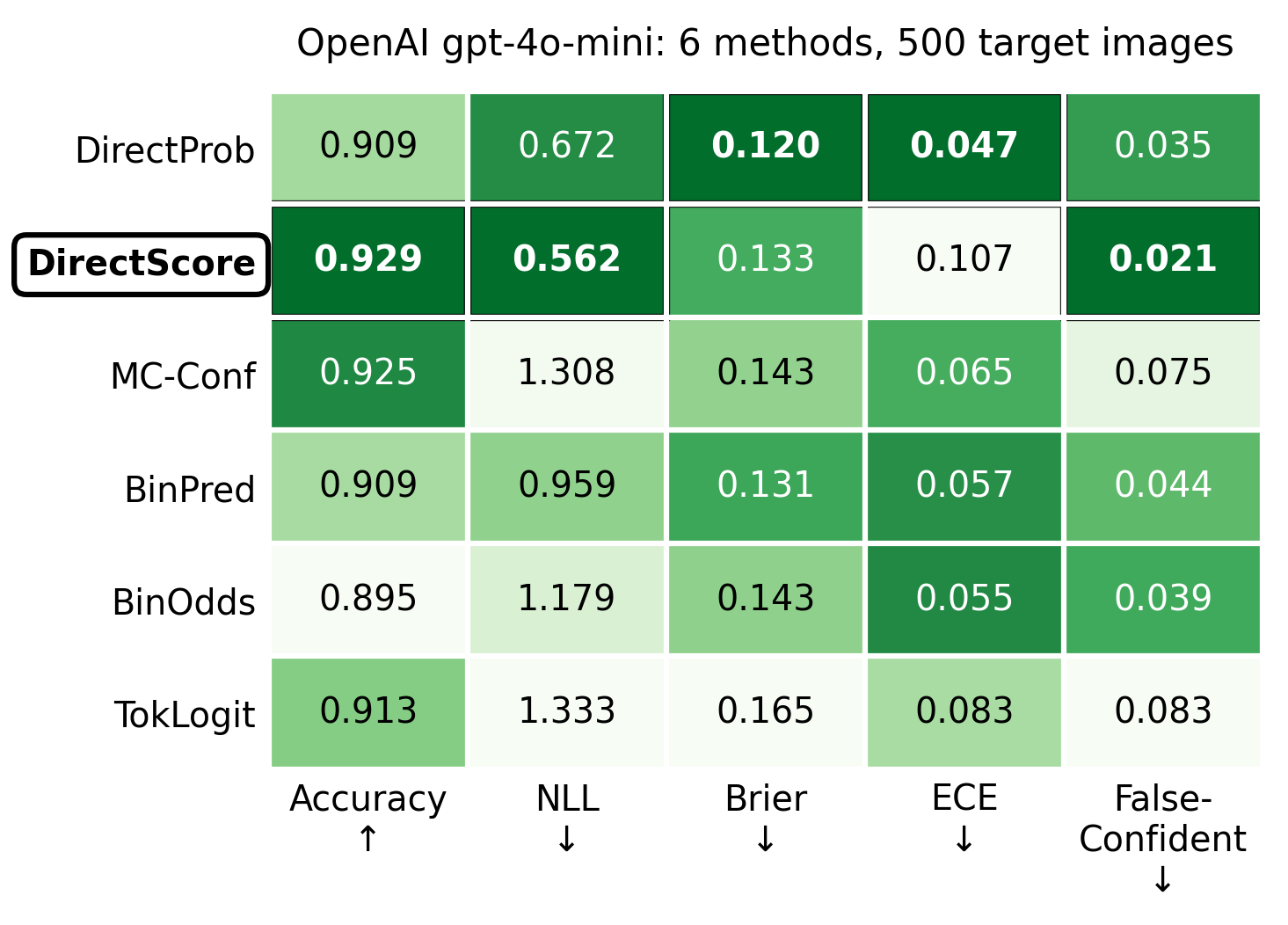}
\caption{Discrimination and calibration across six uncertainty-elicitation interfaces on GPT-4o-mini over 500 target images. DirectScore provides the best overall operating point, achieving the highest accuracy, lowest NLL, and lowest false-confidence rate.}
\label{fig:vlm_methods}
\end{figure}

We evaluate six uncertainty-elicitation interfaces on a controlled benchmark of 500 synthetic targets spanning combinations of fixed categorical visual attributes, including colors such as red, blue, and green; shapes such as triangles, circles, and squares; and surface patterns such as solid, striped, and checkerboard. Each target is evaluated at the attribute level, allowing uncertainty quality to be measured using classification accuracy, negative log-likelihood (NLL), Brier score, expected calibration error (ECE), and false-confidence rate. As shown in~\Cref{fig:vlm_methods}, classification accuracy varies only modestly across interfaces ($0.89$--$0.93$), whereas their uncertainty characteristics differ substantially. 

Direct-Score provides the best operating point for our downstream task, achieving the highest accuracy ($0.929$), lowest NLL ($0.562$), and lowest false-confidence rate ($0.021$), while Direct-Probability achieves the best Brier score ($0.120$) and ECE ($0.047$). Although their overall performance is comparable, Direct-Score is attractive for embodied use because it elicits relative evidence over attribute values and converts it to a normalized categorical distribution explicitly, rather than relying on the VLM to report well-calibrated absolute probabilities. Together with its lower false-confidence rate, which is particularly important for confidence-triggered stopping, we therefore adopt Direct-Score for subsequent experiments.

\subsection{Search-and-Identify Evaluation}
\label{sec:search_experiment}
\begin{figure}[t]
\centering
\includegraphics[width=0.8\linewidth]{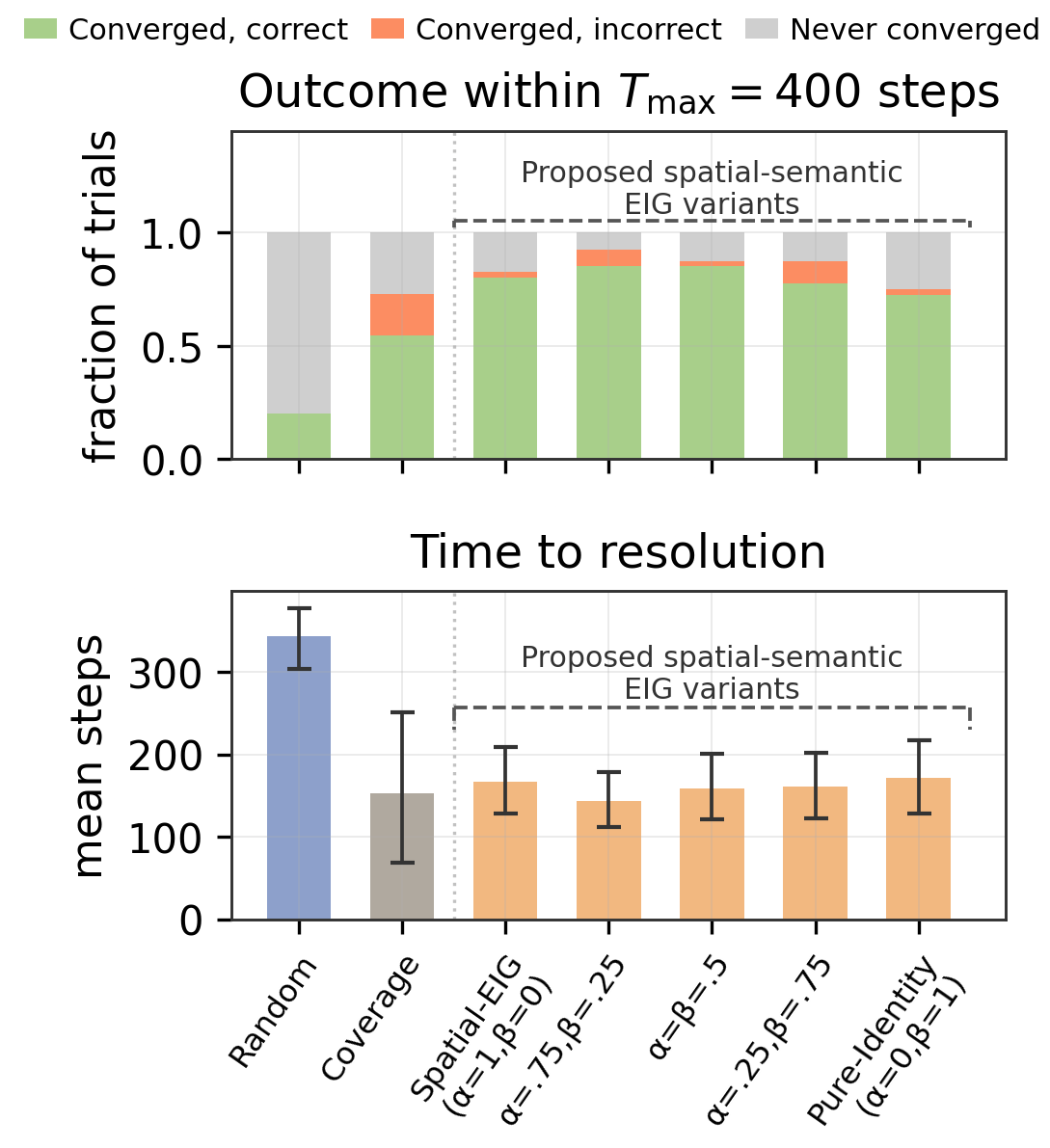}
\caption{Search-and-identify outcomes under blur with confidence-triggered stopping across seven conditions. Top: outcome breakdown. Bottom: mean time to resolution with 95\% bootstrap CI; non-converged trials are counted at
$T_{\max}=400$.}
\label{fig:longrun_all}
\end{figure}

The robot searches a bounded $20\times20$ grid workspace containing $N=10$ static candidate objects. Each environment contains exactly one desired target and hard distractors constructed to share a subset of its target-defining attributes while differing on at least one attribute specified by $\mathcal{D}$. Consequently, no individual attribute is sufficient for identification, and the robot must reason over their joint probabilistic evidence.

To model variable sensing quality, every VLM query is independently degraded using Gaussian blur sampled over $[0,100]\%$, with semantic evidence weighted according to the calibration procedure in~\Cref{sec:semantic_identity_belief}. We compare Random, deterministic Coverage, and five fixed spatial--semantic EIG weightings spanning $(\alpha,\beta)\in\{(1,0),(.75,.25),(.5,.5), (.25,.75),(0,1)\}$. Each condition is evaluated on paired randomized environments using horizon $H=5$. A mission terminates when $\max_j\pi_t(c_j)\geq0.95$, while $T_{\max}=400$ serves only as a safety ceiling.

\subsection{Search Performance} \label{sec:search_results} \Cref{fig:longrun_all} shows that explicitly reasoning about target-identity uncertainty improves search-to-identification performance. EIG-based planners reach a confident decision in $75.0\%$--$92.5\%$ of trials, compared with $20.0\%$ for Random and $72.7\%$ for Coverage. The balanced planner $(\alpha=\beta=0.5)$ correctly converges to the desired target in $85.0\%$ of all trials, a $65$ percentage-point improvement over Random, while identification accuracy among converged EIG trials remains high ($88.6\%$--$97.1\%$). 

The decomposition is particularly useful because semantic EIG makes uncertainty over the identity of $\mathcal{X}$ an explicit information-gathering objective. As semantic weight increases, the robot resolves the target using progressively less exploration: the mean number of candidates discovered before commitment decreases from $8.0$ for pure Spatial-EIG to $6.0$ for Pure-Identity, and the mean number of VLM queries decreases from $16.2$ to $12.9$. Thus, semantic uncertainty derived from the VLM allows the planner to focus sensing on observations that disambiguate the desired target, whereas spatial EIG continues to value discovery of unresolved regions. Mixed spatial--semantic objectives therefore provide a direct mechanism for accelerating target identification without discarding the spatial exploration needed to discover informative candidates.

\section{Conclusion} We introduce a spatial--semantic uncertainty formulation for VLM-based target search that separates uncertainty over candidate locations from uncertainty over target identity. Probabilistic VLM evidence forms a global identity posterior with an undiscovered-target hypothesis, allowing spatial and semantic EIG to separately value candidate discovery and target disambiguation. Experiments show that semantic EIG reduces unnecessary exploration and VLM queries while maintaining high identification accuracy once confidence is reached, whereas spatial EIG favors broader environmental discovery. Thus, the decomposition provides an explicit mechanism for directing information gathering toward target identification or search-space exploration according to mission priorities. The present formulation assumes a known candidate count $N$, controlled synthetic evaluation rather than physical deployment, and a fixed categorical attribute set rather than attributes derived automatically from arbitrary natural language. These assumptions provide a controlled setting for studying spatial--semantic uncertainty. Future work will relax them in real-world and multi-robot settings, where quantified semantic uncertainty can guide active sensing, selective VLM querying, and information sharing.

\newpage
\bibliographystyle{ieeetr}
\bibliography{references}
\end{document}